\documentclass[final]{cvpr}

\usepackage{times}
\usepackage{epsfig}
\usepackage{graphicx}
\usepackage{amsmath}
\usepackage{amssymb}
\usepackage[utf8]{inputenc} 
\usepackage[T1]{fontenc}    
\usepackage[dvipsnames]{xcolor}
\usepackage{graphicx}
\usepackage{soul}
\graphicspath{ {./images/} }
\usepackage{url}            
\usepackage{booktabs}       
\usepackage{amsfonts}       
\usepackage{nicefrac}       
\usepackage{microtype}      
\usepackage{multirow}
\usepackage{mathtools}
\usepackage{floatrow}
\usepackage{wrapfig}
\usepackage{float}
\restylefloat{table}

\usepackage{listings}
\newcommand{\VFactor}{\ensuremath{r_v}}
\newcommand{\QKFactor}{\ensuremath{r_{qk}}}
\newcommand{\BottleneckFactor}{\ensuremath{r_b}}
\newcommand{\BlockSize}{\ensuremath{b}}
\newcommand{\HaloSize}{\ensuremath{h}}
\newcommand{\ImageSize}{\ensuremath{s}}
\newcommand{\ChannelSize}{\ensuremath{c}}
\newcommand{\WindowSize}{\ensuremath{w}}
\newcommand{\BaseWidthFactor}{\ensuremath{r_w}}
\newcommand{\ThirdGroupLayers}{\ensuremath{l_3}}
\newcommand{\FinalFCWidth}{\ensuremath{d_{f}}}
\newcommand{\Heads}{\ensuremath{f}}

\usepackage[pagebackref=true,breaklinks=true,colorlinks,bookmarks=false]{hyperref}
\usepackage{cleveref}

\def\cvprPaperID{10649} 
\def\confYear{CVPR 2021}

\begin{document}

\title{Scaling Local Self-Attention for Parameter Efficient Visual Backbones}

%

\author{%
  Ashish Vaswani \\
  Google Research \\
  \and 
  Prajit Ramachandran \\
  Google Research \\
  \and
  Aravind Srinivas \\
  UC Berkeley \\
  \and 
  Niki Parmar \\
  Google Research \\
  \and
  Blake Hechtman \\
  Google Research \\
  \and
  Jonathon Shlens \\
  Google Research \\
}

\maketitle
\begin{abstract}
    
Self-attention has the promise of improving computer vision systems due to parameter-independent scaling of receptive fields and content-dependent interactions, in contrast to parameter-dependent scaling and content-independent interactions of convolutions.
Self-attention models have recently been shown to have encouraging improvements on accuracy-parameter trade-offs compared to baseline convolutional models such as ResNet-50.
In this work, we aim to develop self-attention models that can outperform not just the canonical baseline models, but even the high-performing convolutional models.
We propose two extensions to self-attention that, in conjunction with a more efficient implementation of self-attention, improve the speed, memory usage, and accuracy of these models.
We leverage these improvements to develop a new self-attention model family, \emph{HaloNets}, which reach state-of-the-art accuracies on the parameter-limited setting of the ImageNet classification benchmark. In preliminary transfer learning experiments, we find that HaloNet models outperform much larger models and have better inference performance. On harder tasks such as object detection and instance segmentation, our simple local self-attention and convolutional hybrids show improvements over very strong baselines. These results mark another step in demonstrating the efficacy of self-attention models on settings traditionally dominated by convolutional models.
\end{abstract}

\section{Introduction}
Vision and natural language processing (NLP) systems divide the landscape of computational primitives. 
While self-attention is the primary workhorse in NLP, convolutions are ubiquitous in nearly all vision models.
Convolutions embody the principle of \emph{local} processing, to learn local spatial features such as edges and texture that are abundant in images.
On the other hand, the Transformer~\cite{vaswani2017attention} showed that self-attention is an effective and computationally efficient mechanism for capturing \emph{global} interactions between words in a sentence.
The success of self-attention in NLP motivates research in understanding how self-attention can improve vision.
Self-attention has several properties that make it a good fit for vision:
(a) content-based interactions as opposed to content-independent interactions of convolution;
(b) parameter-independent scaling of receptive field size as opposed to parameter-dependent scaling of convolution;
(c) empirical ability to capture long-range dependencies for use in larger images;
(d) flexibility to handle and integrate multiple types of data that appear in vision, such as pixels~\cite{wang2018non, bello2019attention, ramachandran2017searching, zhao2020exploring}, point clouds~\cite{yang2019modeling}, sequence conditioning information~\cite{xu2015show}, and graphs~\cite{li2019relation}.
Self-attention may also be regarded as an adaptive nonlinearity paralleling a long history of nonlinear processing techniques in computer vision, such as bilateral filtering \cite{paris2009bilateral} and non-local means \cite{buades2005non}.



Several recent papers~\cite{bello2019attention, ramachandran2019standalone, dosovitskiy2020image, zhao2020exploring, srinivas2021bottleneck} have attempted using self-attention primitives to improve image classification accuracy over the strong and commonly used ResNet backbones~\cite{he2016deep, he2016identity}.
Among them, the Stand-Alone Self-Attention (SASA)~\cite{ramachandran2019standalone} is a fully self-attentive model that replaces every spatial convolution with \emph{local} self-attention, which improves the performance of ResNet backbones while having fewer parameters and floating point operations.
While conceptually promising, these models lag behind state-of-the-art convolutional models in image classification.
State-of-the-art convolutional models \cite{tan2019efficientnet, zoph2018learning, radosavovic2020designing} use a variety of scaling techniques to achieve strong performance across a range of computation and parameter regimes.
In this work, we aim to develop and understand techniques for scaling \emph{local} self-attention models to outperform some of the best convolutional models.
Scaling self-attention models presents a unique set of challenges.
For example, convolutions have been very efficiently mapped to matrix accelerators such as TPUs and GPUs that drive most deep learning workloads, but fast implementations of local 2D self-attention do not currently exist.
To bridge this gap, we introduce a \emph{non-centered} version of local attention that efficiently maps to existing hardware with \emph{haloing}.
While our formulation breaks \emph{translational equivariance}, it improves both throughput and accuracies over the centered local self-attention used in SASA.
We also introduce a strided self-attentive downsampling operation for multi-scale feature extraction.

We leverage these techniques to develop a new local self-attention model family, \emph{HaloNet}, which achieves state-of-the-art performance across different parameter regimes. 
The largest HaloNet achieves 84.9\% top-1 accuracy on the ImageNet~\cite{russakovsky2015imagenet} classification benchmark (Section~\ref{subsec:halonet-sota}). We perform a detailed study to uncover how self-attention and convolutional models scale differently. Our self-attention layers also show promising results on harder tasks such as object detection and instance segmentation (Section~\ref{sec:detection}) using the Mask R-CNN framework on the COCO benchmark. Finally, we end with a discussion of current limitations and ideas for future work in applying self-attention to vision. 

\section{Models and Methods}

\newcommand{\PixelIJ}{\ensuremath{(i, j)}}
\newcommand{\LocalWindow}{\ensuremath{\mathcal{N}\PixelIJ}}

Although our models use self-attention instead of convolutions for capturing spatial interactions between pixels, they adopt some important architectural features of modern convolutional neural networks (CNNs). 
Like CNNs, we compute \emph{multi-scale feature hierarchies}~\cite{lin2017feature} which enable detecting objects at multiple sizes in tasks such as localization and instance segmentation. 
For this, we develop a strided self-attention layer, a natural extension of strided convolutions (Section~\ref{sec:improved-impl}). 
To deal with the computational cost in larger resolutions where global attention is infeasible, we follow the fairly general principle of \emph{local processing}, which is at the heart of convolutions and natural perceptual systems~\cite{hubel1963shape, hubel1968receptive}, and use spatially restricted forms of self-attention. 
However, unlike the model of~\cite{ramachandran2019standalone}, that also use local self-attention, we abstain from enforcing translation equivariance in lieu of better hardware utilization, which improves the speed-accuracy tradeoff (Section~\ref{sec:improved-impl}). 
Also note that while we use local attention, our receptive fields per pixel are quite large (up to $18 \times 18$) and we show in Section~\ref{sec:image_scaling} that larger receptive fields help with larger images. 
In the remainder of this section, we will motivate self-attention for vision tasks and describe how we relax translational equivariance to efficiently map local self-attention to hardware.
\begin{figure*}
    \includegraphics[width=\textwidth]
    {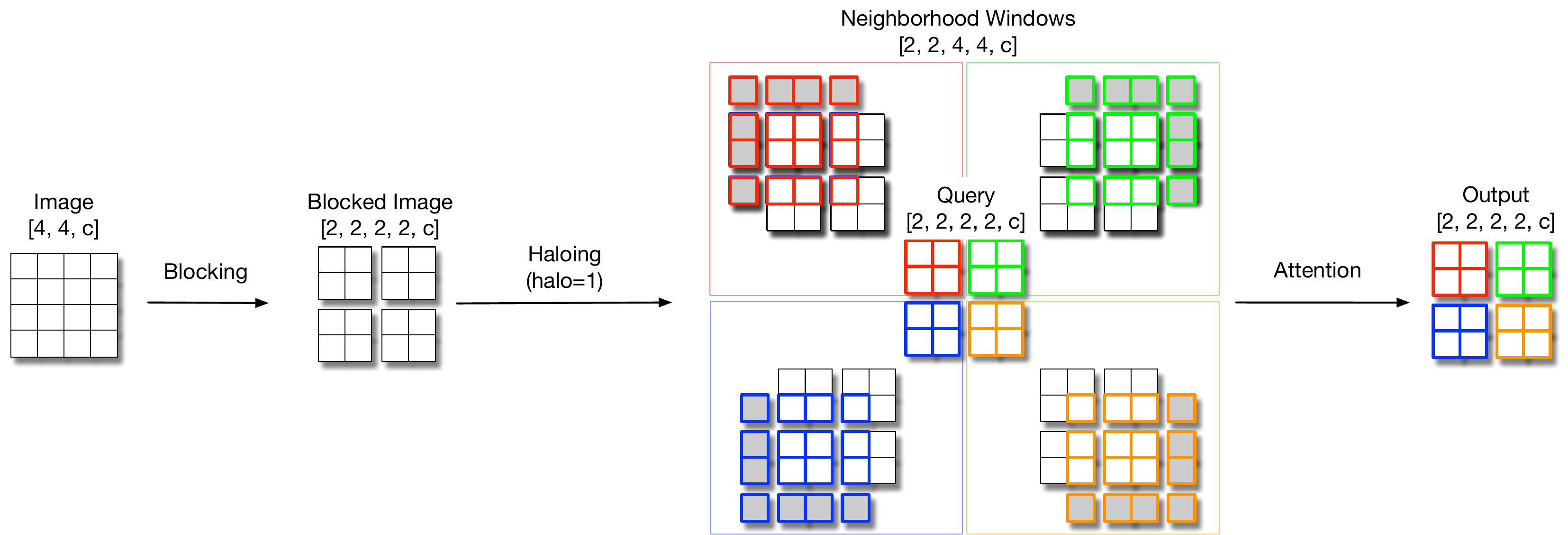} 
  \caption{\textbf{HaloNet local self-attention architecture: The different stages of blocked local attention for a $[4, 4, c]$ image, block size $\BlockSize=2$, and halo $\HaloSize=1$}. The image is first blocked into non-overlapping $[2, 2, c]$ images from which the queries are computed. The subsequent haloing step then extracts a $[4, 4, c]$ memory around each of the blocks which linearly transform to keys and values. The spatial dimensions after attention are the same as the queries.}
  \label{fig:haloing-steps}
\end{figure*}

\begin{figure}
    \includegraphics[width=\textwidth]
    {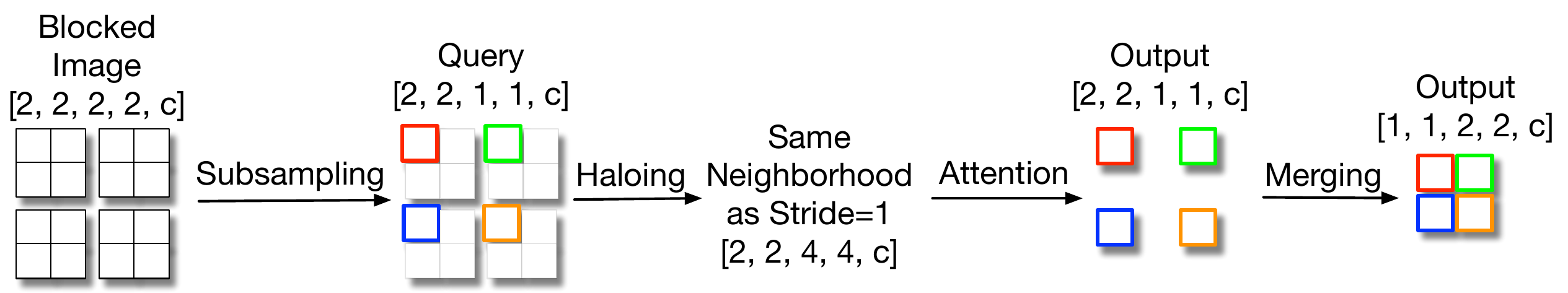} 
  \caption{The attention downsampling layer subsamples the queries but keeps the neighborhood the same as the the stride=1 case.}
  \label{fig:striding}
\end{figure}

\subsection{Self-attention can generate spatially varying convolutional filters}
\label{sec:methods:background}
Recent work~\cite{cordonnier2019relationship} has shown that self-attention with sufficient number of heads and the right geometric biases can simulate convolutions, suggesting a deeper relationship between self-attention and convolutions.  Self-attention has been viewed as a method to directly capture relationships between distant pixels~\cite{ramachandran2019standalone, hu2019local, wang2020axial}. 
It has also been interpreted as a specific instantiation of the classic technique of non-local means \cite{buades2005non,wang2018non}.
The perspective that we discuss in this section is one that views  self-attention as generating spatially varying filters, in contrast to the reuse of the same filter across every spatial location in standard convolutions~\cite{elsayed2020revisiting}. 
To observe this, we write self-attention and convolution as specific instances of a general spatial pooling function. 
Given an input $x \in \mathcal{R}^{H \times W \times c_{in}}$, where $H$ is the height, $W$ is the width, and $c_{in}$ is the number of input channels, we define a local 2D pooling function that computes an output at location $\PixelIJ$, $y_{ij} \in \mathcal{R}^{c_{out}}$ as

\[y_{ij} = \sum_{a, b \in \LocalWindow}  f(i, j, a, b) x_{ab}, \]

where $f(i, j, a, b)$ is a function that returns a weight matrix $W \in \mathcal{R}^{c_{in} \times c_{out}}$ at every location in a 2D window $\LocalWindow$ of size $k \times k$ centered at $\PixelIJ$.
Note that later in this section, we introduce non-centered windows for self-attention, but we use centering here for ease of explanation.
This computation is repeated for every pixel $\PixelIJ$.
For a convolution, $f(i, j, a, b)$ returns a \emph{different linear transformation} for each relative distance in neighborhood, and these weights are shared across all $\PixelIJ$.
Weight sharing significantly reduces parameters and encourages learning features that repeat spatially.
In dot-product relative self-attention~\cite{shaw2018self,ramachandran2019standalone, bello2019attention} (\cref{eq:self-att-func-1,eq:self-att-func-2}), every pixel in the neighborhood shares the \emph{same linear transformation} which is multiplied by a scalar probability that is a function of both content-content and content-geometry interactions resulting in weights that can vary spatially.
As an example, for a ball and an orange at two different locations in an image, pixels inside the ball and the orange are likely to generate different $p_{a-i, b-j}^{i j}$ because of the different content around them, such as color or texture.
\begin{align}
f(i, j, a, b)^{conv} & =  W_{a-i, b-j}  \label{eq:conv} \\
\begin{split}
f(i, j, a, b)^{self-att}  & = \texttt{softmax}_{a b}\Big((W_Q x_{i j})^\top W_K x_{a b} +  \\
 & \quad (W_Q x_{i j})^\top r_{a-i,b-j} \Big) W_V  \label{eq:self-att-func-1}
\end{split} \\
 & = p_{a-i, b-j}^{i j} W_v \label{eq:self-att-func-2}
\end{align}

For self-attention, $W_Q$, $W_K$, and $W_V$ are learned linear transformations that are shared across all spatial locations, and respectively produce \emph{queries}, \emph{keys}, and \emph{values} when used to transform $x$. 
Spatial geometry is captured by $r_{a-i,b-j}$, which is a learned relative position based embedding.
The $(W_Q x_{i j})^\top W_K x_{a b}$ component captures the content-content interaction between the query pixel and a key pixel in the window.
The $(W_Q x_{i j})^\top r_{a-i,b-j}$ component is the content-geometry interaction that captures the relationship between the query and the relative position of the key pixel \cite{shaw2018self}. 
Note that this formulation preserves \emph{translational equivariance}.
If an object translates in an image, for any pixel within the object, the content around it stays the same, generating the same $p_{a-i, b-j}^{i j}$, thereby producing the same output after self-attention.
To increase expressivity, multi-headed attention \cite{vaswani2017attention} is used, which repeats this computation multiple times in parallel with different parameters, analogous to group convolutions~\cite{krizhevsky2012, xie2017aggregated}.  

In the \emph{SASA} model of \cite{ramachandran2019standalone}, the local window $\LocalWindow$ is a $k\times k$ window centered around $\PixelIJ$, just like a convolution.
The size of this local window $k$ is an important setting to leverage in self-attention.
Unlike dense convolutions, $k$ can grow without significantly increasing the number of parameters.
Since the projection parameters ($W_Q$, $W_K$, $W_V$) are independent of $k$, the only parameters that increase with $k$ is $r_{a-i,b-j}$.
However, $r_{a-i,b-j}$ constitutes a trivial fraction of the parameters compared to the projection parameters
\footnote{
For a window size as large as $63$, and $16$ dimensions per attention head, $r_{a-i,b-j}$ would add only $63*16=1008$ parameters per layer because $r_{a-i,b-j}$ are shared among heads. 
In contrast, if the dimensions of the attention layer were $512$, $W_Q$, $W_K$, $W_V$ would contribute $786432$ parameters. We show details in the appendix.}
, so increasing $k$ does not not impact the number of parameters of the layer significantly.
In contrast, the number of parameters in a convolution layer scale quadratically with $k$ (\eg, a $5\times5$ convolution has $\frac{25}{9}$ times the parameters of a $3\times3$ convolution).
On the other hand, the computational cost of self-attention grows quadratically with $k$, preventing the use of very large values for $k$.

\subsection{Improving the speed-memory tradeoff by relaxing translational equivariance}\label{sec:improved-impl}
\emph{Global} self-attention, in which all locations attend to each other, is too expensive for most image scales due to the quadratic computation cost with respect to $k$.
Thus, multi-scale visual backbones need to use local attention to limit the size of $k$.
We follow the intuitive form of local attention developed in \cite{ramachandran2019standalone}, which tries to mimic the square neighborhoods used by convolutions.
This form of local attention requires extracting local 2D grids around each pixel.
Unfortunately, while deep learning libraries automatically handle neighborhood gathering for convolutions, no such neighborhood gathering function exists for local self-attention (or any general local function).
Thus, implementing local self-attention requires explicitly gathering the local neighborhoods before the actual self-attention operation can be performed.
While the implementation of this local neighborhood gathering function might initially appear to be a relatively minor implementation detail, in practice, it must actually be carefully designed to reduce memory usage while avoiding unnecessary extra computation.
An unoptimized implementation can prevent self-attention models from scaling up due to either out-of-memory errors or excessive slowness.
The following discussion frames the design considerations of this neighborhood gathering function.

\begin{table}[tb]
\resizebox{\columnwidth}{!}{
\begin{tabular}{c c c c}
\toprule
\textbf{Method}       & 
\textbf{\begin{tabular}[c]{@{}c@{}}Neighborhood\\ Memory\end{tabular}} & \textbf{\begin{tabular}[c]{@{}c@{}}Receptive\\ Field\end{tabular}}                                                               &       
\textbf{\begin{tabular}[c]{@{}c@{}}FLOPs\\ Per Pixel\end{tabular}}  \\
\midrule
Global   & $HW\ChannelSize $ $HW$                                 & $4HW \ChannelSize  $                                 \\
Per pixel windows   & $HWk^2\ChannelSize$                                              & $k \times k$                                                 & $4k^2\ChannelSize$                                                                                                               \\
SASA~\cite{ramachandran2019standalone} & $\frac{HW}{\BlockSize^2}
(\BlockSize+2\HaloSize)^2 \ChannelSize$ & $k \times k$, where $\HaloSize=\lfloor \frac{k}{2} \rfloor$ & $4 ( \BlockSize + 2\HaloSize)^2 \ChannelSize$                                                                                                              \\
Blocked local (ours) & $\frac{HW}{\BlockSize^2}
(\BlockSize+2\HaloSize)^2 \ChannelSize$ & $( \BlockSize + 2\HaloSize) \times ( \BlockSize + 2\HaloSize)$ & $4( \BlockSize + 2\HaloSize)^2 \ChannelSize$                                                              \\
\bottomrule
\end{tabular}} 
\caption{\textbf{Scaling behavior of self-attention mechanisms}. $\Heads$ is the number of heads, $\BlockSize$ is the size of the block, $c$ is the total number of channels, and $\HaloSize$ is the size of the halo}\label{tab:complexities}
\end{table}

A straightforward approach would gather $k \times k$ sized windows \emph{separately} around each pixel.
As summarized in Table~\ref{tab:complexities} (Row 1), this method  blows up the memory used by a factor of $k^2$ due to replicating the pixel contents for each of the $k^2$ neighborhoods it participates in.
This solution quickly leads to out-of-memory errors.
\emph{Global} attention (Row 4) is at the other end of the spectrum, where all pixels \emph{share} the same neighborhood, lowering memory at the expense of considerably more FLOPs \footnote{To illustrate this, on a $128 \times 128$ resolution with $64$ channels, global self-attention would incur about $28$ times more FLOPs than a $3 \times 3$ convolution with $64$ input and output channels}.
This solution slows down models significantly, while also imposing memory problems due the massize size of the attention matrix.
A solution that lies in-between these two extremes should trade-off memory and compute appropriately, with the recognition that a small amount of waste is required.

A compromise solution can be achieved by leveraging the idea that \emph{neighboring pixels share most of their neighborhood}.
For example, two pixels that are right next to each other share $k \times (k-1)$ pixels of their neighborhoods.
Thus a local neighborhood for a \emph{block} of pixels can be extracted once together, instead of extracting separate neighborhoods per pixel.
The FLOPs can be controlled by varying the number of pixels that form a block.
We name this strategy \emph{blocked local self-attention}.
The two extremes discussed above are a special case of blocked local self-attention.
Global attention corresponds to setting the block size to be the entire spatial extent, while the per-pixel extraction corresponds to setting the block size to be 1.

Figure~\ref{fig:haloing-steps} depicts the different steps involved in executing blocked local self-attention for an image with height $H=4$, width $W=4$,
and $\ChannelSize$ channels with stride $1$.
Blocking chops up the image into a $\frac{H}{\BlockSize}, \frac{W}{\BlockSize}$ tensor of \emph{non-overlapping} $(\BlockSize, \BlockSize)$ blocks.
Each block behaves as a group of query pixels and a \emph{haloing} operation combines a band of $\HaloSize$ pixels around them (with padding at boundaries) to obtain the corresponding \emph{shared neighborhood} block of shape $(\frac{H}{\BlockSize}, \frac{W}{\BlockSize}, \BlockSize+2\HaloSize, \BlockSize+2\HaloSize, \ChannelSize)$ from which the keys and values are computed.
$\frac{H}{b} \times \frac{W}{b}$ attention operations then run in parallel for each of the query blocks and their corresponding neighborhoods, illustrated with different colors in Figure~\ref{fig:haloing-steps}.
SASA~\cite{ramachandran2019standalone} used the same blocking strategy\footnote{Code for both SASA and HaloNet will be made available, along with the checkpoints for HaloNet}, setting $\HaloSize=\lfloor \frac{k}{2} \rfloor$ and uses attention masks to emulate pixel-centered neighborhood windows of size $k \times k$.
For example, to achieve a $7 \times 7$ pixel centered window, \cite{ramachandran2019standalone} set $\HaloSize=3$.
The use of attention masks gives the operation translational equivariance, since each pixel only looks at a square window around it. 

However, the downside of using attention masks is that it wastes computation that must happen regardless due to the implementation of this algorithm.
If attention masks are not used, the receptive field increases without any additional computation, as shown in Table~\ref{tab:complexities} (Rows 2 and 3).
However, pixel-level translational equivariance is lost because the non-square receptive fields means that the output of a pixel is dependent on which block it falls into.
Take for example a pixel at the left edge of its block, which sees additional pixels that are to the right of its square receptive field.
If the entire image is shifted one pixel to the right, the pixel now falls into right edge of a neighboring block, and now sees additional pixels that are to the left of its square receptive field.
Thus the output of the pixel is dependent on its position in a block, which can change if the image shifts.
Another perspective is that blocked local self-attention is only translational equivariant to shifts of size $b$.
While pixel-level translational equivariance is considered important for achieving good performance\cite{zhang2019making}, we find that empirically, using a non-masked block local self-attention actually improves the accuracy of the model (see Section \ref{sec:relaxing-translation-equivariance}).
We suspect that the image shifting and cropping perturbations in common data augmentation strategies reduce the reliance on such inductive biases. 
Thus we adopt unmasked blocked local self-attention because it improves accuracy without sacrificing performance.

\begin{figure}[h]
    \includegraphics[width=0.9\textwidth]{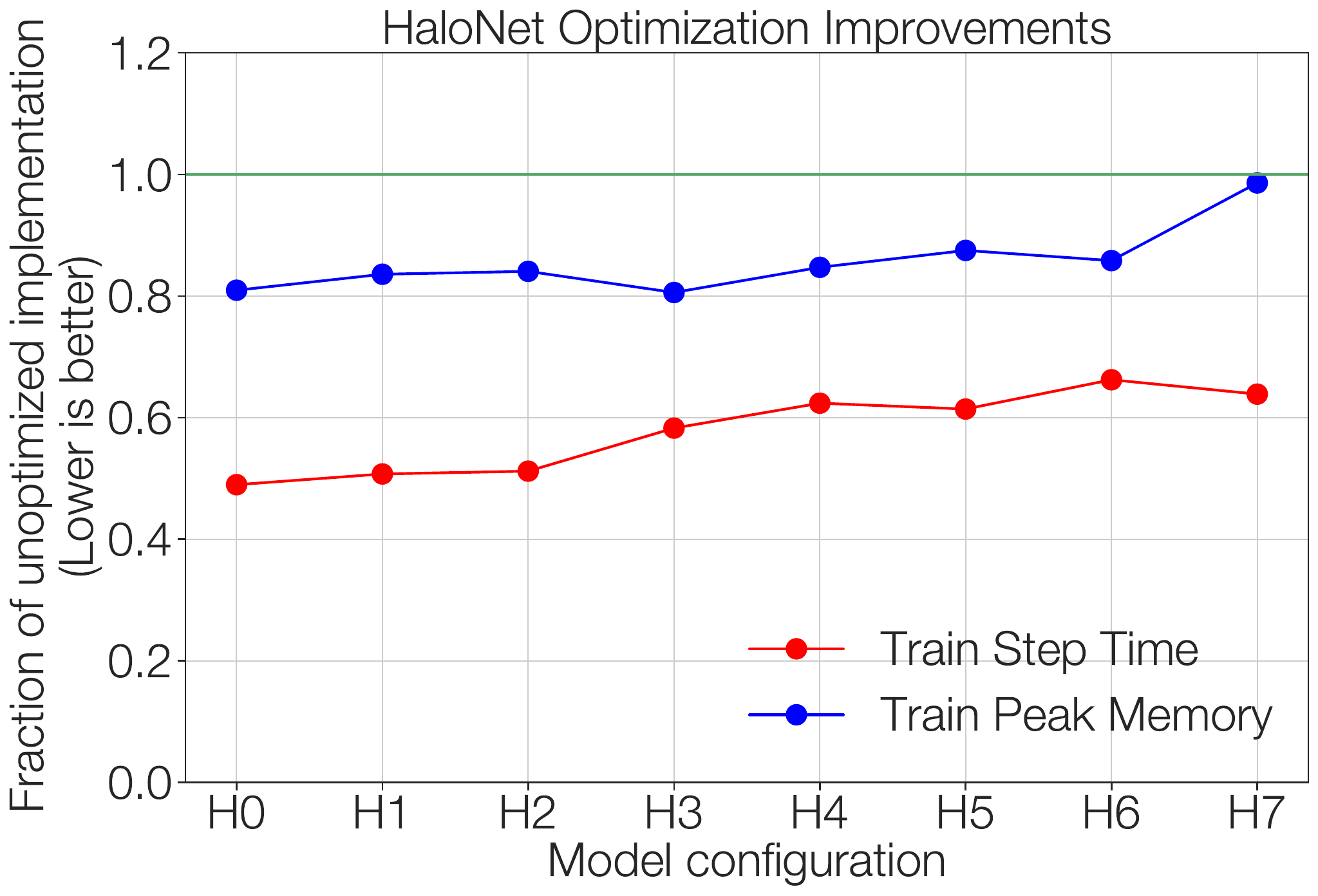}%
  \caption{\textbf{Optimizations improve performance.} The improvements here are a result of reducing FLOPs with our attention downsampling and improved local self-attention algorithms that avoid reshapes and data formatting. In some cases, we halve the training step time computed on TPU v3.}%
  \label{fig:experiments:speed-memory-improvements}
\end{figure}

Another difference with SASA is our implementation of downsampling.
We replace attention followed by post-attention strided average pooling by a single strided attention layer that subsamples queries similar to strided convolutions, as shown in Figure~\ref{fig:striding}. Note that we use the same neighborhood as is extracted in the stride $1$ case (Figure~\ref{fig:haloing-steps}). This change does not impact accuracy while also reducing the FLOPs $4\times$ in the downsampling layers. We also implement some important algorithmic optimizations that improve our throughput primarily by avoiding reshapes and data formatting operations. In interest of space, we list them in the Appendix~\ref{sec:optimizations}.  Taken together, the speedups produced by these improvements are significant as seen in Figure~\ref{fig:experiments:speed-memory-improvements}, with up to 2$\times$ improvements in step time.
These improvements can be leveraged to train large self-attention models that were previously too expensive. 
We leave additional optimizations, such as fused operations and better pipelining of memory accesses with computation, to future work.

To conclude this section, it's important to note that in  the deeper layers of multiscale architectures, smaller spatial dimensions and larger channels would shift the compute calculus in favor of global attention. The models we introduce in Section~\ref{subsec:halonet-fam}, also take advantage of this, typically using local attention in the higher resolutions and global attention when the image resolutions are the smallest. 


\newfloatcommand{capbtabbox}{table}[][\FBwidth]
\newcommand{\LayerImageSize}[1]{$\frac{\ImageSize}{#1}\times\frac{\ImageSize}{#1}$}
\newcommand{\LayerStructure}[2]{$\left\{\begin{tabular}[c]{@{}c@{}} $1 \times 1, {#1}$ \\ $\text{attention}(\BlockSize , \HaloSize), {#1} \cdot \VFactor$ \\ $1 \times 1, {#1} \cdot \BottleneckFactor$ \end{tabular}\right\} \times {#2}$}

\capbtabbox{%
\small
\begin{tabular}{cc}
    \toprule
     \textbf{\begin{tabular}[c]{@{}c@{}}Output\\Resolution\end{tabular}}  & \textbf{Layers}                                                                                                                                                                             \\
    \midrule
    \LayerImageSize{4} & \begin{tabular}[c]{@{}c@{}}$7 \times 7$ conv stride 2, 64\\ $3 \times 3$ max pool stride 2\end{tabular}                                                                                   \\
    \midrule
    \LayerImageSize{4} & \LayerStructure{64}{3} \\
    \midrule
    \LayerImageSize{8} & \LayerStructure{128}{3} \\
    \midrule
    \LayerImageSize{16} & \LayerStructure{256}{\ThirdGroupLayers} \\
    \midrule
    \LayerImageSize{32} & \LayerStructure{512}{3} \\
    \midrule
    \LayerImageSize{32} & $1\times1, \FinalFCWidth$ \\
    \midrule
    $1\times1$  & \begin{tabular}[c]{@{}c@{}} global average pooling \\ fc, 1000 \end{tabular} \\
    \bottomrule
\end{tabular}
}{%
  \caption{\textbf{HaloNet model family specification.}}%
\label{table:methods:network-structure}
  
}


\subsection{HaloNet} \label{sec:halonet}
Using the implementation of local 2D self-attention with haloing detailed above, we propose a new model, \emph{HaloNet} that matches state-of-the-art convolutional models on the parameter-accuracy trade-off curve.
We leverage the structure of ResNets \cite{he2016deep} that stack multiple residual bottleneck blocks together (see Table \ref{table:methods:network-structure}). 
HaloNet uses a few minor modifications from ResNets: (a) adding a final $1\times1$ convolution before the global average pooling for larger models, following EfficientNet \cite{tan2019efficientnet}, (b) modifying the bottleneck block width factor, which is traditionally fixed at $4$, (c) modifying the output width multiplier of the spatial operation, which is traditionally fixed at $1$, (d) changing the number of blocks in the third stage from $4$ to $3$ for computational reasons because attention is more expensive in the higher resolution layers. 
We also fix the number of heads for each of the four stages to $(4, 8, 8, 8)$ because heads are more expensive at higher resolutions. To summarize, the scaling dimensions in HaloNet are: image size $\ImageSize$, query block size $\BlockSize$, halo size $\HaloSize$, attention output width multiplier $\VFactor$, bottleneck output width multiplier $\BottleneckFactor$, number of bottleneck blocks in the third group $\ThirdGroupLayers$, and final $1 \times 1$ conv width $\FinalFCWidth$. Our attention neighborhoods range from $14 \times 14$ ($\BlockSize=8, \HaloSize=3$) to $18\times18$ ($\BlockSize=14, \HaloSize=2$). 


Since the ResNet structure was initially designed for convolutions, we suspect that designing architectures specifically for attention may improve HaloNet. In our work we maintained homogeneity across all layers of model for hyperparameters such as the block ($\BlockSize)$ and halo ($\HaloSize$) sizes.  We also hope that using automated architecture search methods~\cite{tan2019efficientnet} to optimize these hyperparameters for specific accelerators will lead to better local attention architectures. In our work, we train with comparable image sizes as EfficientNet models to determine if attention models can scale to larger images. 

\section{Related Work}
\label{sec:related-works}

Attention has steadily risen in adoption in vision models in recent years.
First introduced in various forms of sequence modeling~\cite{graves2013generating,bahdanau2014neural,vaswani2017attention, parikh2016decomposable,lin2017structured}, attention was used to attend to image features in the text generation module of image captioning models~\cite{xu2015show}.
Attention is also closely related to non-local means~\cite{buades2005non}, a pairwise-weighted global sum of pixels originally developed for image denoising. 
\cite{wang2018non} applied non-local means on top of spatially downsampled convolutional features to improve video classification.
However, since these methods scale quadratically with receptive field size, they cannot be used because the spatial size is too large. In order to apply self-attention, \cite{parmar2018image} applies local attention on images for the task of image generation. \cite{bello2019attention} spatially downsample the features for attention and concatenate the attention outputs to convolutional features.
Instead, we directly build on top of the approach of~\cite{ramachandran2019standalone}, who compute attention on local regions in order to build a fully self-attentional vision model for classification and object detection.
Different forms of attention for pure self-attention vision models have also been proposed~\cite{hu2019local,zhao2020exploring}, which are orthogonal and complementary to the focus on scaling in this work.
In addition to attention over the spatial extent that we focus on, components that perform attention over channels have also been used to augment convolutional models~\cite{hu2018squeeze,li2019selective}. In recent and concurrent work, Vision Transformer~\cite{dosovitskiy2020image} show that applying transformers on projections of \emph{non-overlapping} image patches can achieve accuracies comparable to SOTA when pre-trained on very large (JFT-300M~\cite{sun2017revisiting}) and medium sized (ImageNet-21k~\cite{imagenet_cvpr09}) classification datasets. However, their models do not adopt a multiscale architecture and our focus in this work is training on ImageNet~\cite{russakovsky2015imagenet} from scratch. In Section~\ref{sec:imagenet-21k-transfer}, we conduct transfer experiments and compare with ViT and BiT~\cite{kolesnikov2019big}.

Generally, the performance of computational primitives tend to improve over time due to algorithmic changes to the primitive and better software implementations.
In particular, convolution have improved over the last decade through changes in
(a) the computation of the primitive~\cite{chellapilla2006high,jia2014caffe,mathieu2013fast,vasilache2014fast,winograd1980arithmetic,lavin2016fast};
(b) the software implementation~\cite{chetlur2014cudnn};
(c) the structure of the primitive itself, through for example, grouped convolution~\cite{xie2017aggregated} and depthwise separable convolution~\cite{sifre2014rigid}.
Attention is in the beginning phases of this performance improvement trajectory, and given its importance in sequence modeling~\cite{vaswani2017attention}, it will likely see sustained effort to enhance performance.
Local attention could also receive performance improvements if it is adopted more widely to combat the general problem of processing large inputs.
Our work introduces blocked local attention to efficiently process immediate neighbors.
Other forms of non-global pixel interaction can also be implemented efficiently \cite{child2019generating, ho2019axial, wang2020axial, bello2021lambdanetworks}.

\section{Experiments}
\label{subsec:halonet-fam}

Each HaloNet model (H0--H7) is designed by successively growing the values of the hyperparameters defined in Table~\ref{table:methods:network-structure}.
In interest of space, we leave the exact configurations of our models to the Appendix~\ref{sec:detailed-models}. We also leave the training and evaluation of larger HaloNet models that compare with larger EfficientNet models for future work.

\begin{figure}[]
    \centering
    \includegraphics[width=1.0\textwidth]{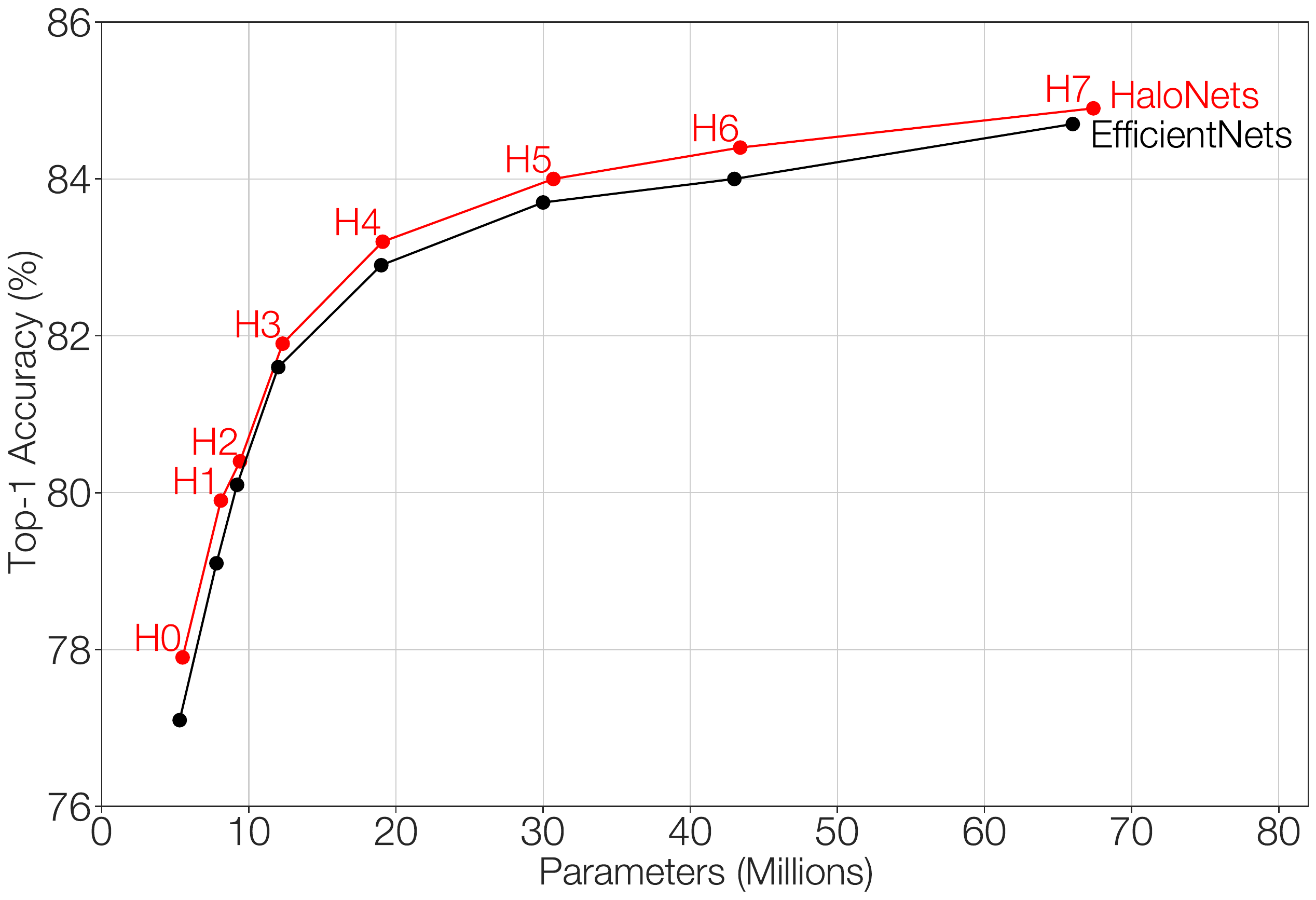}    
    \caption{\textbf{HaloNets can match EfficientNets on the accuracy vs. parameter trade-off}. The accuracies for EfficientNets B5 and B7 were obtained using RandAugment.} 
    \label{fig:experiments:halo-efficient-param-tradeoff}
\end{figure}

\subsection{HaloNets are competitive with state-of-the-art convolutional models}
\label{subsec:halonet-sota}
We train our HaloNet models on ImageNet~\cite{russakovsky2015imagenet} (ILSVRC-2012) benchmark with a batch size of $4096$ and learning rate of $1.6$, which is linearly warmed up for 10 epochs and followed by cosine decay~\cite{loshchilov2016sgdr}.
The models are trained for $350$ epochs with Nesterov’s Accelerated Gradient~\cite{nesterov1983,icml2013_sutskever13}, and regularized with dropout~\cite{srivastava2014dropout}, weight decay, RandAugment~\cite{cubuk2019randaugment} and stochastic depth~\cite{huang2016deep}.

We find that HaloNets perform at par or slightly better (Figure~\ref{fig:experiments:halo-efficient-param-tradeoff}) than EfficientNet models for the same parameters, outperforming other model families. Our best model, H7, achieves \textbf{84.9}\% top-1 ImageNet validation accuracy and \textbf{74.7}\% top-1 accuracy on ImageNet V2~\cite{recht2019imagenet} (with a -0.5\% gap to the linear fit in~\cite{recht2019imagenet}). For each of our HaloNet models, we use image sizes comparable to the corresponding EfficientNet model, training on images sizes up to $600\times600$. (Table~\ref{tab:appendix:model-description}).
For a comparison of our latencies with EfficientNet, the reader can refer to Section~\ref{sec:performance}.
To the best of our knowledge, these results are the first to show that self-attention based models for vision perform on par with the SOTA for image classification when trained on imagenet from scratch. Note that for all our experiments, we report accuracies at the end of training and we tune regularization hyperparameters such as augmentation hyperparameters for the baselines and HaloNet models. 

\subsection{Model study 1: comparing self-attention and convolutions}
\label{sec:experiments:studies:attention-vs-conv}

In the following sections, we will focus on model studies to distinguish the advantages of self-attention over convolutions for vision and and understand how to best design self-attention vision architectures.
This knowledge is important since much of the progress in convolutional networks comes from improvements in architecture design while keeping the core convolution primitive the same~\cite{krizhevsky2012,szegedy2016rethinking,he2016deep}. We believe our study is the first to explicitly examine the design of optimal self-attention vision architectures.

For the remainder of the experimental section, we compare with ResNet-50~\cite{he2016identity}, the canonical vision model, because many of the components that we ablate have been well studied for ResNet-50, allowing us to use best practices for the baseline model.
We tune our baseline ResNet-50 implementation to achieve a better accuracy, 77.6\%, compared to commonly reported numbers in the literature. For example, ~\cite{he2016deep} report 76.3\%. We then create a new HaloNet architecture, HaloNet-50, that exactly matches the ResNet-50 architecture by replacing spatial convolutions with local self-attention.
HaloNet-50 and ResNet-50 have about $18.0$ million and $25.5$ million parameters respectively. We train both for $150$ epochs on $256\times256$ image size. We share other training details of the ablation set-up in the appendix 

\subsubsection{Transfer of convolutional components to self-attention}\label{sec:transfer-components}

Utilizing regularizations and architectural modules beyond the core primitive is critical for achieving strong results~\cite{he2019bag}.
In this section, we study the effects of these additional components on self-attention models.
The components we study were all designed for use in convolutional models, as they were developed through experimentation (either human or automated search) on convolutional models.
We examine whether these components can successfully transfer to the new model family of self-attention networks.

\newcommand{\regularization}[1]{{#1}}
\newcommand{\architecturemodule}[1]{{#1}}
We focus on 4 different components based on the design of EfficientNet~\cite{tan2019efficientnet}, 2 \architecturemodule{architecture modules} and 2 \regularization{regularizations}: 
\architecturemodule{Squeeze-and-Excitation (SE)} \cite{hu2018squeeze}, a channel attention module used after the spatial convolution;
\architecturemodule{SiLU/Swish-1} \cite{ramachandran2017searching, elfwing2018sigmoid, hendrycks2016gaussian}, an activation function with the form $x \cdot \text{sigmoid}(x)$;
\regularization{RandAugment (RA)}~\cite{cubuk2019randaugment}, a data augmentation scheme that simplifies AutoAugment~\cite{cubuk2019autoaugment};
and \regularization{Label Smoothing (LS)}~\cite{szegedy2016rethinking}, a smoothing of the label distribution.

The results of adding these various components to the baseline is in Table \ref{table:experiments:component-ablation}.
Suprisingly, regularizations of the same strength improve HaloNet accuracies significantly more than ResNet, despite HaloNet having around 30\% fewer parameters than ResNet.
When label smoothing and RandAugment are added, HaloNet improves by 1.3\% while ResNet improves by 0.8\%.
This result suggests that self-attention models may require regularizations that are typical of larger convolutional models, perhaps due to the expressivity of self-attention.

However, the architecture modules that were developed for convolutional models only improve attention models by a small amount.
When Squeeze-and-Excitation (SE) and SiLU/Swish-1 are added, ResNet improves by 1.3\% while HaloNet only improves by 0.4\%. We speculate that HaloNet models benefit from the gating and multiplicative interactions that comprise self-attention and do not need explicit gating such as SE. Further research must be conducted in order to discover architecture modules that can consistently improve a variety of self-attention models. Inspired by these findings, we decided to use label smoothing, SiLU/Swish-1, and RandAugment in our HaloNet $H0-H7$ models. We also use stochastic depth for our larger models~\cite{huang2016deep, tan2019efficientnet}.

\begin{table}[h]
   \begin{center}
    \resizebox{8cm}{!}{%
    \begin{tabular}{l|cc|cc}
    \toprule
    \multicolumn{1}{l|}{\textbf{Components}} & \textbf{\begin{tabular}[c]{@{}c@{}}HaloNet \\ Accuracy\end{tabular}} & \textbf{\begin{tabular}[c]{@{}c@{}}Baseline\\ $\Delta$\end{tabular}} & \textbf{\begin{tabular}[c]{@{}c@{}}ResNet \\ Accuracy\end{tabular}} & \textbf{\begin{tabular}[c]{@{}c@{}}Baseline\\ $\Delta$\end{tabular}} \\
    \midrule
    Baseline  & 78.6 & 0.0 & 77.6 & 0.0   \\
    \midrule
    + LS & 79.7 & 1.1   & 78.1 & 0.5 \\
    + LS, RA  & 79.9  & 1.3  & 78.4 & 0.8  \\
    + SE  & 78.6  & 0.0  & 78.6  & 1.0 \\
    + SE, SiLU/Sw1  & 79.0 & 0.4 & 78.9 & 1.3 \\
    + LS, SE  & 79.7 & 1.1 & 78.9 & 1.3 \\
    + LS, SE, SiLU/Sw1 & 79.9 & 1.3 & 79.1 & 1.5  \\
    + LS, SE, SiLU/Sw1, RA & 80.5  & 1.9& 79.5  & 1.9\\       
    \bottomrule
    \end{tabular}%
    }

    \end{center}
    \caption{
       \textbf{HaloNet improves more than ResNet with regularizations, but does not improve significantly with architectural modules that strongly benefit ResNet.} 
       Starting from a baseline model, adding \regularization{label smoothing (LS)}, \regularization{RandAugment (RA)}, \architecturemodule{Squeeze-and-Excitation (SE)}, and \architecturemodule{SiLU/Swish-1 (SiLU/Sw1)}.
    }
    \label{table:experiments:component-ablation}
\end{table}


\subsubsection{Increasing image sizes improve accuracies} \label{sec:image_scaling}
\begin{figure}[]
    \centering
        \includegraphics[width=0.8\textwidth, scale=1]{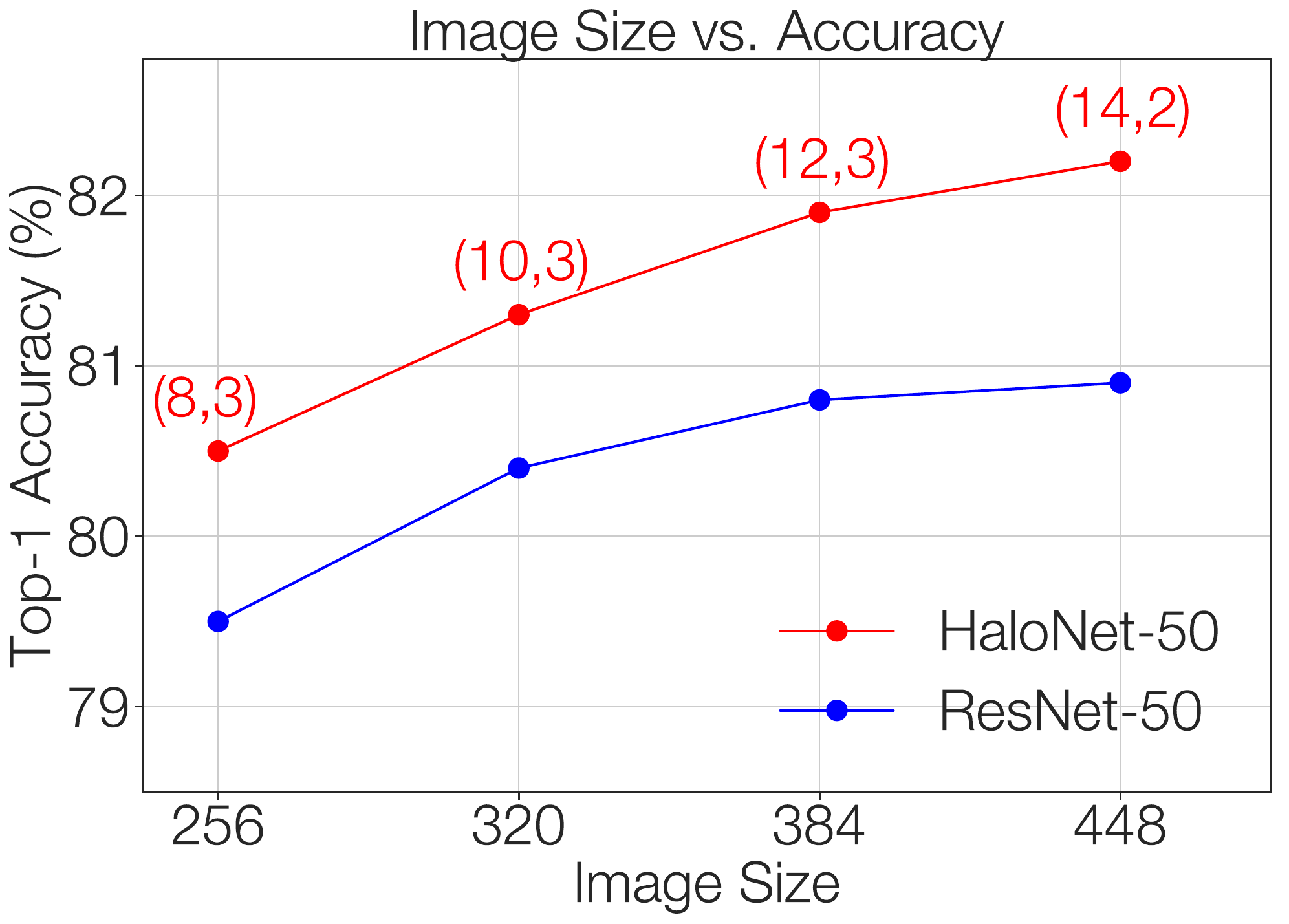}
        \caption{\textbf{The accuracy gap between HaloNet-50 and ResNet-50 is maintained with increasing image sizes.} The HaloNet experiments are annotated with block size ($\BlockSize$), halo size ($\HaloSize$).}
        \label{fig:experiments:subfig:image-scaling}
    \end{figure}

A beneficial property  of self-attention is attention is that the receptive field size can scale along with image size without significantly impacting the number of parameters (see Section~\ref{sec:methods:background}).
As shown in Figure~\ref{fig:experiments:subfig:image-scaling}, HaloNet consistently improves when using larger images. Although we also see improvements with convolutional models, the accuracy gap between HaloNets and ResNets is maintained.

\subsection{Model study 2: HaloNet architecture study}
\label{sec:experiments:studies:halonet-arch}

\begin{figure}[h]
    \centering
    \includegraphics[width=0.8\textwidth]{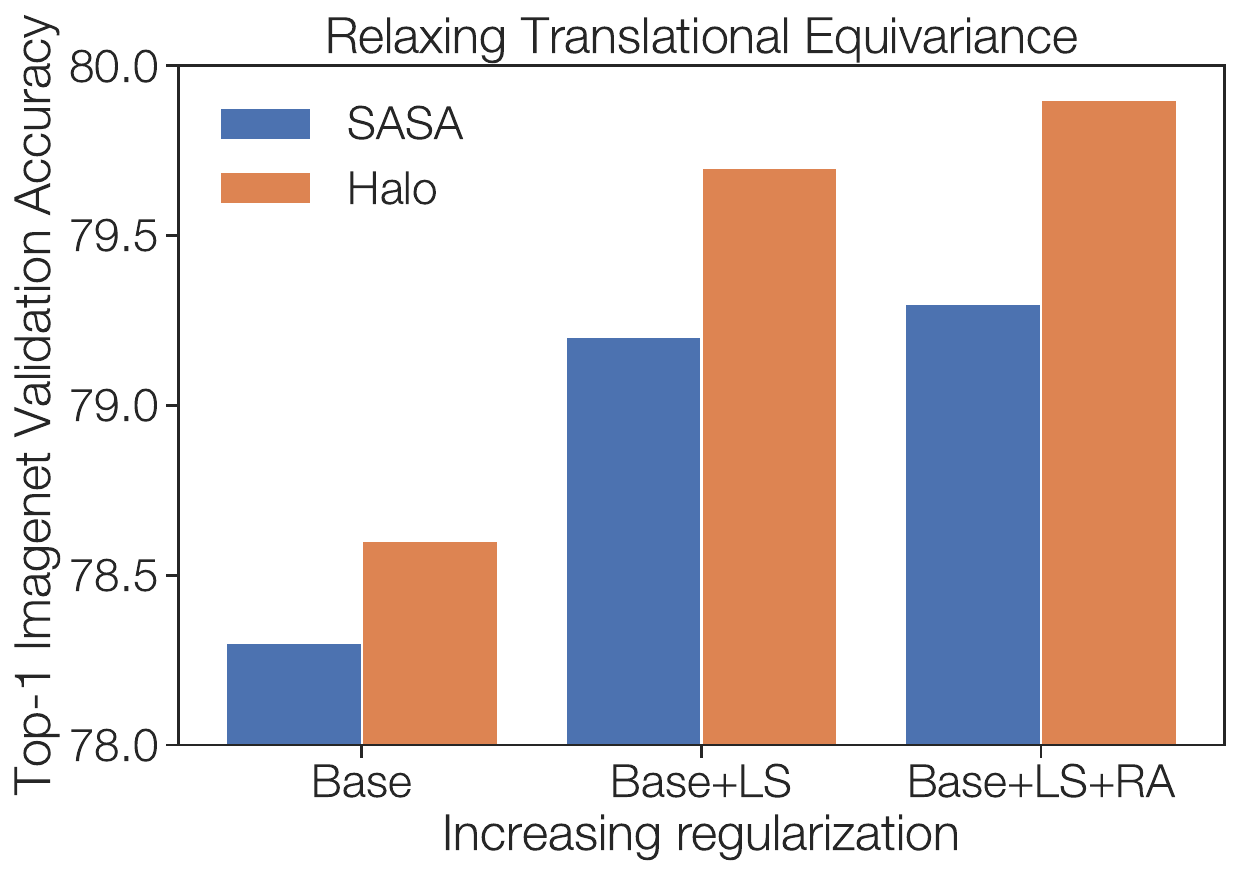}%
  \caption{\textbf{Relaxing translational equivariance improves accuracies}} \label{fig:relaxing-trans-eq}
\end{figure}

In this section, we will study the impact of relaxing translational equivariance and the relationship of neighborhood window and halo sizes. In the interest of space, a detailed study of scaling various components of our models such as $\VFactor$, $\QKFactor$ etc can be found in the Appendix~\ref{subsec:scaling-halonet}.

\paragraph{Relaxing translational equivariance:}
\label{sec:relaxing-translation-equivariance}
In Figure~\ref{fig:relaxing-trans-eq}, we see that HaloNet-50 with $\BlockSize=8$, and $\HaloSize=3$ achieves better accuracies using the same block and halo to achieve $7\times7$ neighborhoods with attention masks ~\cite{ramachandran2019standalone} and the gap widens with more regularizations. This suggests that larger receptive fields are more important than inductive biases such as translational equivariance.

\paragraph{Window and halo size:} When using the blocked input format, there are two ways of changing the window size of attention: changing the query block size or the halo size. For the same window size $\WindowSize$, smaller query blocks and larger halos require more memory than larger query blocks and smaller halos, as discussed in section~\ref{sec:improved-impl}.

\begin{figure}[]
        \includegraphics[width=0.8\textwidth]{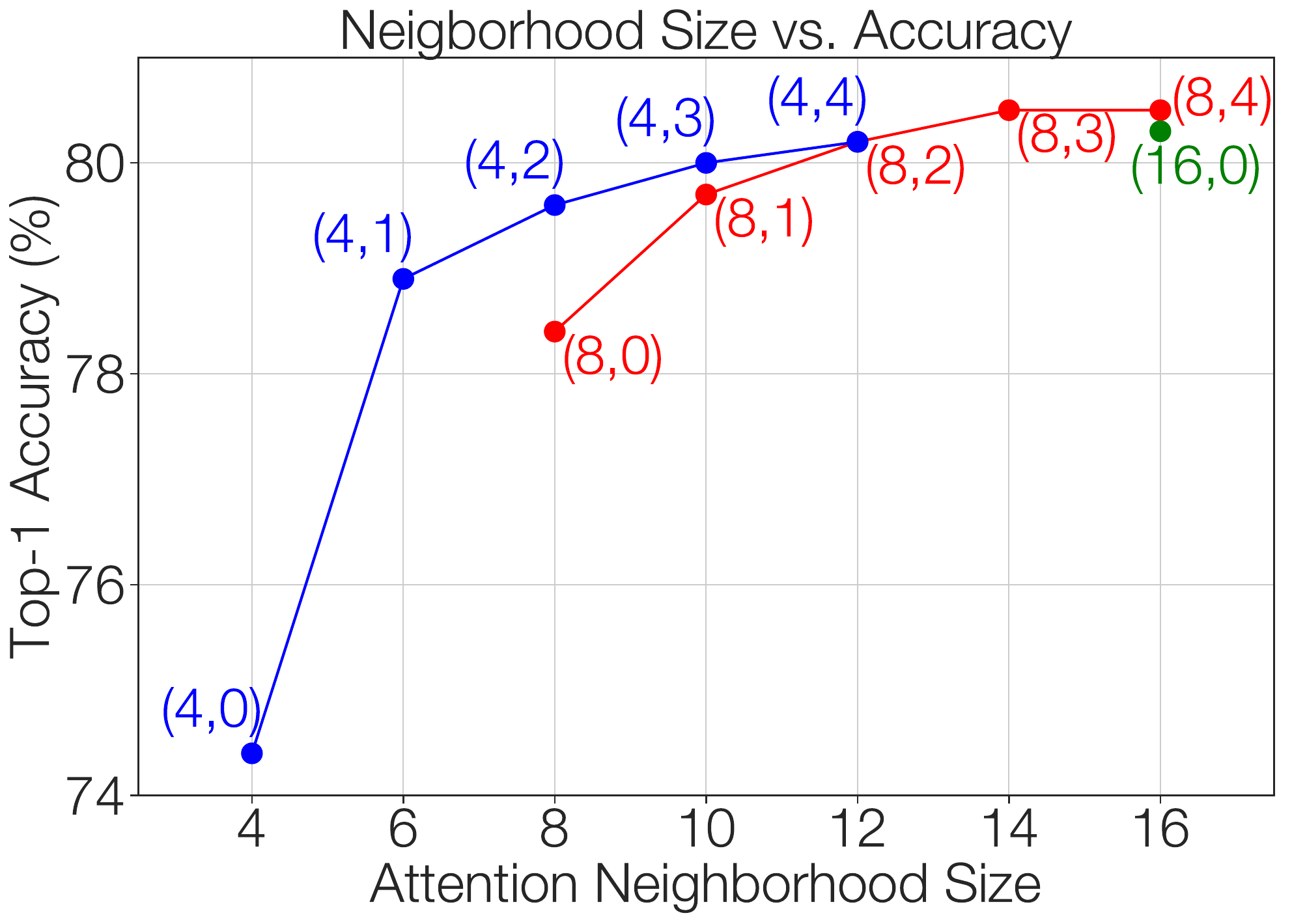}
    \caption{\textbf{Increasing window sizes improves accuracy up to a point.} The experiments in the graph have been annotated with their block size ($\BlockSize$), halo size ($\HaloSize$), $\HaloSize=0$ implies attention with \emph{non-overlapping} blocks}
    \label{fig:experiments:image-scaling-and-block-sizes}
\end{figure}

We see in Figure~\ref{fig:experiments:image-scaling-and-block-sizes} that accuracy consistently improves as the window size increases.
In particular, doubling the window size from $6\times6$ to $12\times12$ produces a $1.3\%$ accuracy gain. These results suggest that increasing window size can be successfully used to scale models without increasing the number of parameters, potentially beneficial for production environments. Furthermore, for a fixed window size, the choice of query block size does not impact results, enabling the usage of larger query block sizes to reduce memory.
Figure~\ref{fig:experiments:image-scaling-and-block-sizes} also shows that eschewing haloing for  \emph{non-overlapping} attention, can lower accuracy significantly unless the blocks are quite large. For example using a block size of $4$ and a halo of $1$ results in better accuracy than using a block size of $8$ with $0$ halo, despite a smaller neighborhood size.

\subsection{Convolution-Attention hybrids improve the speed-accuracy tradeoff}
In our final set of ablations, we replace self-attention with convolutions to understand where attention layers are currently most beneficial. In Table~\ref{tab:hybrids}, we show results for replacing attention layers with convolutions with squeeze-and-excitation modules in each of the stages of our best performing model (HaloNet H7). Having convolutions in all stages except the last yields the fastest model albeit with a significant loss in top-1 accuracy (1\%). Splitting the allocation between convolutions (in stages 1--2) and attention (in stages 3–4) minimally detriments predictive accuracy while significantly improving training and inference step times. We leave a detailed study of improved hybrid models for future work.

\begin{table}[]
\resizebox{6.5cm}{!}{%
\begin{tabular}{c c c c}
\toprule
\textbf{\begin{tabular}[c]{@{}c@{}}Conv\\ Stages\end{tabular}} & \textbf{\begin{tabular}[c]{@{}c@{}}Attention\\ Stages\end{tabular}} & \textbf{\begin{tabular}[c]{@{}c@{}}Top-1\\ Acc (\%)\end{tabular}} & \textbf{\begin{tabular}[c]{@{}c@{}}Norm.\\ Train\\ Time\end{tabular}} \\
\midrule 
-  & 1, 2, 3, 4 & 84.9 & 1.9   \\
1 & 2, 3, 4  & 84.6  & 1.4  \\
1, 2  & 3, 4  & 84.7 & 1.0   \\
1, 2, 3  & 4 & 83.8 & 0.5   \\                  \bottomrule                           
\end{tabular} 
}
\caption{\textbf{Replacing attention layers with convolutions in stages 1 and 2 exhibit the best speed vs. accuracy tradeoff.} All the models had about $67$ million parameters and the train and inference times are normalized to the corresponding times for EfficientNet B7. Please see Figure~\ref{fig:discussion:speed-of-halonet-vs-efficientnet} for a comparison of step time with other HaloNet models.}\label{tab:hybrids}
\end{table}

\begin{table*}[h]
\small
\centering
\resizebox{17.25cm}{!}{
\begin{tabular}{c c c c c c c c c c c}
\toprule
\textbf{Model}  & $\mathbf{AP^{bb}}$ & $\mathbf{AP^{bb}_s}$ & $\mathbf{AP^{bb}_m}$ & $\mathbf{AP^{bb}_l}$ & $\mathbf{AP^{mk}}$ & $\mathbf{AP^{mk}_s}$ & $\mathbf{AP^{mk}_m}$ & $\mathbf{AP^{mk}_l}$ &  \textbf{\begin{tabular}[c]{@{}c@{}}Speed \\  (ms)\end{tabular}} & \textbf{\begin{tabular}[c]{@{}c@{}}Train\\ time \\  (hrs)\end{tabular}}\\
\midrule\
R50 baseline in lit  & 42.1 & 22.5 & 44.8 & 59.1 & 37.7 & 18.3 & 40.5 & 54.9 & 409 & 14.6\\
\midrule\
R50 + SE (our baseline)          & 44.5 (+2.4)     & 25.5        & 47.7        & 61.2        & 39.6 (+1.9)     & 20.4        & 42.6        & 57.6    & 446   &  15.2 \\
R50 + SE + Local Att ($\BlockSize=8$)  & 45.2 (++0.7)     & 25.4        & 48.1        & 63.3        & 40.3 (++0.7)      & 20.5        & 43.1        & 59.0   & 540  &  15.8  \\
R50 + SE + Local Att ($\BlockSize=32$)  & 45.4 (++0.9)     & 25.9        & 48.2        & 63.0        & 40.5 (++0.9)      & 21.2        & 43.5        & 58.8   & 613  &  16.5  \\
\midrule
R101 + SE (our baseline)          & 45.9 (+3.8)     & 25.8        & 49.5        & 62.9        & 40.6 (+2.9)     & 20.9        & 43.7        & 58.7    & 740    &   17.9  \\
R101 + SE + Local Att ($\BlockSize=8)$ & 46.8 (++0.9)     & 26.3        & 50.0        & 64.5        & 41.2 (++0.6)     & 21.4        & 44.3   & 59.8    & 799  & 18.4 \\
\bottomrule
\end{tabular}
}

\caption{
\textbf{Accuracies on object detection and instance segmentation.}
We experiment with two settings for self-attention in the last stage: A block size of ($\BlockSize$) of $8$ and a halo size ($\HaloSize$) of $3$ and also with ($\BlockSize=32, \HaloSize=3$) for ResNet-50.
$bb$ (bounding box) refers to detection, and $mk$ (mask) refers to segmentation.
The identifiers $s$, $m$, and $l$ refer to small, medium, and large objects respectively.
Speed is measured as the milliseconds taken by only the backbone (and not the FPN) for a batch size of $32$ on $2$ TPUv3 cores. The train time the total training time calculated from the peak images/sec of the Mask-RCNN training run on 8 TPUv3 cores with a batch size of 64. 
}
\label{tab:detection-detailed}
\end{table*}

\begin{table*}[h]
\centering
\resizebox{17cm}{!}{
\begin{tabular}{c c c c c c c}
\toprule
\textbf{Model}     & \textbf{\begin{tabular}[c]{@{}c@{}}Parameters\\ (Millions)\end{tabular} } & \textbf{ \begin{tabular}[c]{@{}c@{}}Pretraining\\ Image Size\\ (Pixels)\end{tabular} }& \textbf{\begin{tabular}[c]{@{}c@{}}Pretraining\\ Step Time \\(32 per core)\end{tabular} }& \textbf{\begin{tabular}[c]{@{}c@{}}Finetuning\\ Image Size\end{tabular} }& \textbf{\begin{tabular}[c]{@{}c@{}}Finetuning\\ Top-1\\ Accuracy (\%)\end{tabular}} & \textbf{\begin{tabular}[c]{@{}c@{}}Inference\\ Speed\\ img/sec/core\end{tabular} }\\
\midrule
H4 (base 128) & 85                                                              & 256                                                                         & 377 ms                                                                & 384/512                                                         & 85.6/85.8                                                                 & 121.3/48.6                                                                   \\
H4 (base 128, $4\times4$ patch) & 85                                                              & 256                                                                         & 366 ms                                                                & 384/512                                                         & 85.4/85.4                                                                 & 125.7/56.5                                                                   \\
H4 (base 128, Conv-12) & 87                                                              & 256                                                                         & 213 ms                                                                & 384/512                                                         & 85.5/85.8                                                                 & 257.6/120.2                                                                   \\
ViT-L/16   & 300                                                             & 224                                                                         & 445 ms                                                                & 384/512                                                         & 85.2/85.3                                                                 & 74.6/27.4                                                                    \\
BiT-M      & 928                                                              & 224                                                                         &  1021 ms                                                               & 384                                                             & 85.4                                                                      &    54.2                                                                   \\
\bottomrule
\end{tabular}
}
\caption{\textbf{HaloNet models pretrained on ImagetNet-21k perform well when finetuned on ImageNet}. For HaloNet and ViT, we finetuned on $384 \times 384$ and $512 \times 512$ size images. The pretraining step time reports the TPUv3 compute time for a batch size of 32 per core. The inference speed is also computed on a single TPUv3 core.}
\label{tab:imagenet21k-acc}
\end{table*}

\subsection{Transfer from ImageNet-21k}
\label{sec:imagenet-21k-transfer}
Our experiments thus far have focused on training from scratch on ImageNet-ILSVRC-2012~\cite{russakovsky2015imagenet}, where regularizations and longer training are critical for good accuracies. Papers such ~\cite{dosovitskiy2020image, kolesnikov2019big} have shown that a short finetuning step after pretraining models on larger labelled datasets such as ImageNet-21k~\cite{imagenet_cvpr09} or JFT-300M~\cite{sun2017revisiting} can achieve better accuracies without the need for regularization. To understand the transfer properties of HaloNet models, we scale up HaloNet-H4 by increasing the base width to $128$ and evaluate the transfer protocol from~\cite{kolesnikov2019big}, pretraining on the public ImageNet-21k dataset, and finetuning on ImageNet. Following our observation in Table~\ref{tab:hybrids}, we train a hybrid version of this model with convolutions in the first two stages. Note that our hybrids can be seen as using a series of convolution layers to downsample the image. Since we  For a fair comparison with~\cite{kolesnikov2019big}, we do not use squeeze-and-excitation~\cite{hu2018squeeze} in the stages with convolutions. We also investigate the effect of replacing the convolutional stem with linear projections of $4\times4$ non-overlapping patches, similar to the vision Transformer. The details of the models can be found in the Appendix~\ref{sec:appendix:imagenet-21k-transfer}. ImageNet-21k contains $14.2$ million annotated images, and 21k labels, both an order of magnitude larger than ImageNet. Following~\cite{kolesnikov2019big}, we pretrain for $90$ epochs with a batch size of $4096$, and a base learning rate of $0.16$, which is linearly warmed up for $2$ epochs followed by cosine decay~\cite{loshchilov2016sgdr}. We also use a weight decay of $0.00008$, and train with Nesterov’s Accelerated Gradient~\cite{nesterov1983,icml2013_sutskever13} during pretraining and finetuning. We pretrain on $256 \times 256$ size images and finetune on different image sizes, as shown in Table~\ref{tab:imagenet21k-acc}. Our wider H4 and hybrid-H4 models achieves better accuracy than the Vision Transformer and a $4\times$ wide ResNet-152 from~\cite{kolesnikov2019big} and are also faster at inference on larger images. We finetune for $8$ epochs on ImageNet, initializing with the parameters learned from pretraining except for the label embedding matrix, which is initialized to zeros. We train with a batch size of $512$, a learning rate of $0.016$ and cosine decay after linearly warming it up for $0.5$ epochs. We benefit from finetuning with a label smoothing of $0.1$ during finetuning despite pretrainig on a larger dataset. We do not use Polyak averaging~\cite{polyak1992acceleration}, and other regulariations during finetuning. 

Our preliminary results on transfer are promising since we achieve better parameter-accuracy and speed-accuracy tradeoffs than other models on this dataset. We leave the study of transfer with larger HaloNet and HaloNet hybrids for future work. The speed advantages of our models on larger images make them desirable for challenging structured prediction tasks on large images such as object detection and instance segmentation, which we briefly explore in the next section.

\subsection{Detection and instance segmentation}\label{sec:detection}

To understand if our primitives will generalize to structured prediction tasks on larger images, we conduct initial investigations with the simple attention-convolutional hybrids on detection and instance segmentation, using the Mask R-CNN~\cite{he2017mask} framework. These hybrids are also faster and consume less memory than pure attention models, enabling faster experimental cycles. We only replace the last 3 convolutional layers in the ResNet-50 and ResNet-101 backbones with two halo layers with block size, $\BlockSize=8$ and halo size $\HaloSize=3$ (Rows 3 and 6 in Table~\ref{tab:detection-detailed}). For ResNet-50, we also examine using $\BlockSize=32$ and halo size $\HaloSize=3$ to understand benefits from larger receptive fields.  We also use squeeze-and-excitation with convolutions and pre-train them on $512\times512$ images with the regularizations mentioned in Section~\ref{sec:transfer-components}: label smoothing, RandAugment, and stochastic depth. We train our models on the COCO dataset~\cite{lin2014microsoft} with $1024\times1024$ size images for $32$ epochs, using the Cloud TPU Detection Codebase \footnote{ \url{https://github.com/tensorflow/tpu/tree/master/models/official/detection}}. We provide more training details in the Appendix~\ref{sec:detection-appendix}.

Our ResNet-50 baseline in row 2 of Table~\ref{tab:detection-detailed}, is significantly better than what is usually reported in the literature (row 1). Our attention variants achieve at least 0.7 mAP gains on bounding box detection and at least 0.6 mAP gains on instance segmentation on top of our stronger baselines (denoted by \textbf{++} in rows 3, 4 and 6 in Table~\ref{tab:detection-detailed}). 
The gain from local attention with block size $\BlockSize=8$ closes half of the mAP gap between the R50 and R101 baselines in detection and 70\% of the gap in instance segmentation despite being less than a third of the gap in terms of wall-clock time.
Local attention with $\BlockSize=8$ and $\HaloSize=3$ also improves on top of the deep R101 backbone.
Interestingly, localization of large objects ($AP^*_l$) shows the largest improvement when attention is used.
Larger block sizes ($\BlockSize=32$ in row 4) achieve very close performance to $\BlockSize=8$ while being slower. However, we see that $\BlockSize=32$ does much better than $\BlockSize=8$ on small objects ($AP^{*}_s$). 
Future work can combine the best of these two settings. Note that with $\BlockSize=32$, the last two attention layers do global attention since the image is downsampled to $\frac{1024}{32} = 32$ pixels in each spatial dimension. Concurrent work, BoTNet~\cite{srinivas2021bottleneck}, uses global self-attention in ResNet-Attention hybrids for structured prediction tasks and classification. See~\cite{srinivas2021bottleneck} for additional details on the efficacy of global attention for localization tasks 

These models have only three layers of self-attention, and more layers could alter these results. 
We leave the study of detection and instance segmentation with pure attention models to future work.

\section{Discussion}
\label{sec:performance}

\begin{figure}
    \includegraphics[width=0.9\textwidth]{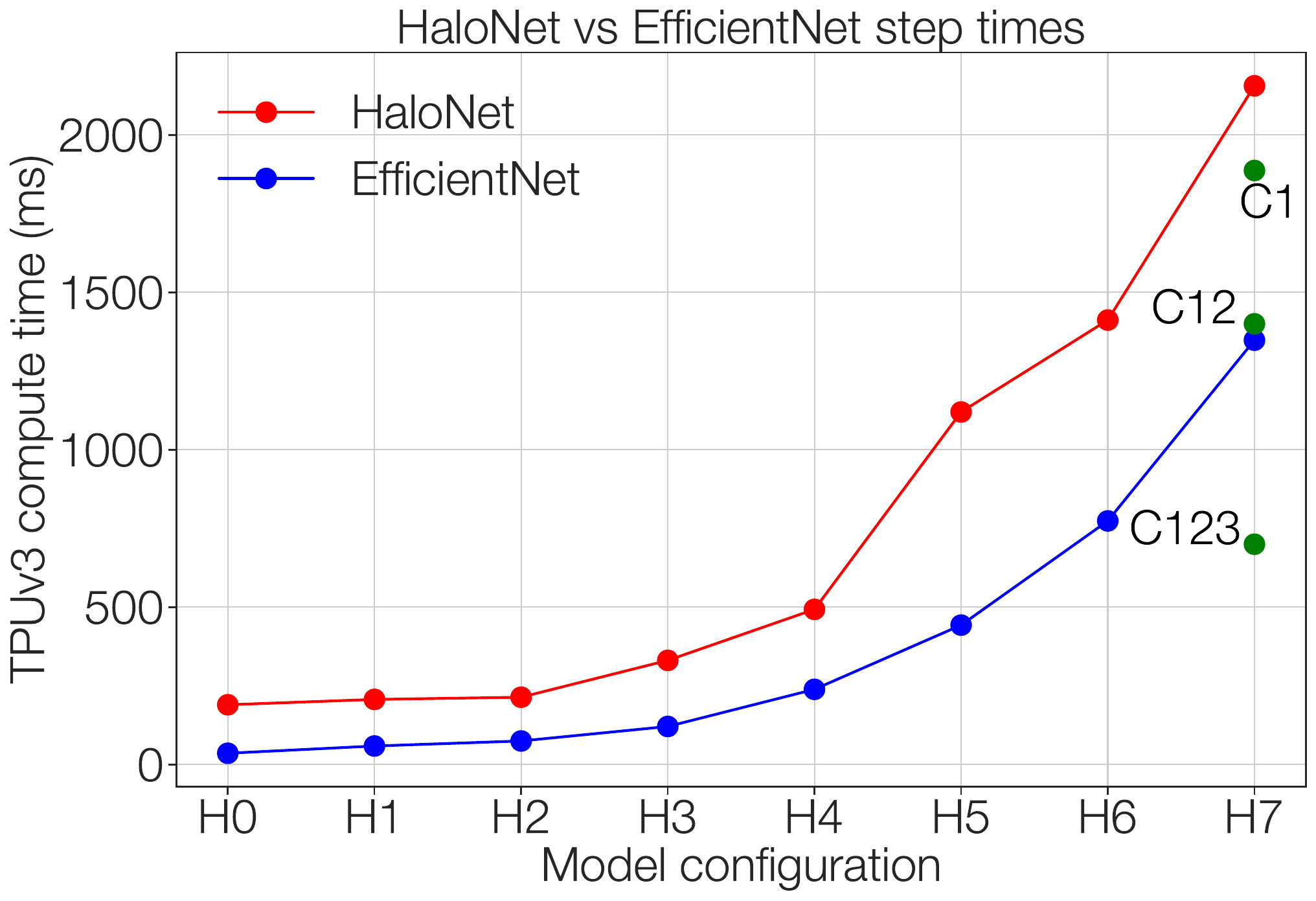}
    \caption{
        \textbf{Pure attention based HaloNet models are currently slower to train than efficient net models.} 
        The times are the TPUv3 compute time needed to process a batch size of $32$ per core. The points in green with annotations C1, C12, and C123 correspond to the hybrid models with convolutions in stages 1, 1--2 and 1--3 respectively. (see Table~\ref{tab:hybrids}).
    }
    \label{fig:discussion:speed-of-halonet-vs-efficientnet}
\end{figure}

In this work, we built multiscale self-attention models that are competitive with the best convolutional models.
To achieve this result, we developed two attention improvements: blocked local attention and attention downsampling.
We also performed multiple ablations to understand how to improve the scaling of self-attention models. 

Our results demonstrate that self-attention is competitive accuracy-wise when training on ImageNet from scratch. 
Figure~\ref{fig:discussion:speed-of-halonet-vs-efficientnet} shows that pure self-attention\footnote{By pure attention we mean models that use self-attention in all layers except the stem, which is convolutional.} based HaloNets are currently slower to train than the corresponding EfficientNets and require further optimizations for large batch training. However, our hybrids have the same speed-accuracy tradeoff as EfficientNets. On transfer from ImageNet-21k, our models outperform very strong models such as BiT~\cite{kolesnikov2019big} and ViT~\cite{dosovitskiy2020image}, on both accuracy and speed.
Model optimizations, such as using architecture search methods to find better speed-accuracy tradeoffs or different forms of more powerful and/or efficient attention forms \cite{zhao2020exploring,roy2020efficient}, are promising directions for machine learning researchers.
Implementation optimizations, such as better memory management, can improve the practicality of these models. Also, scaling up our models to larger widths might cause our operations to transition from being memory bound to compute bound, and lead to better speed-accuracy tradeoffs. We leave this study for future work. Overall, our work shows that self-attention can be competitive in regimes traditionally dominated by computer vision.
Future work can  push these boundaries further, both in terms of scale and efficiency.
\section{Acknowledgements}
We would like to thank David Fleet for valuable discussions. We would also like to thank Irwan Bello, Barret Zoph, Mingxing Tan, and Lucas Beyer for valuable commentary on earlier drafts of the paper.



{\small
\bibliographystyle{ieee_fullname}
\bibliography{main}
}

\clearpage
\section*{Appendix}
\renewcommand{\thetable}{A\arabic{table}}
\setcounter{table}{0}
\renewcommand{\thefigure}{A\arabic{figure}}
\setcounter{figure}{0}
\renewcommand{\thesection}{\Alph{section}}
\setcounter{section}{0}


\section{Relative embeddings add very few parameters to the model}
Our parameters grow very slowly with receptive field. In this section, we will show that the number of parameters in the relative embeddings, the only spatially dependent parameters, is quite small. 
As described in the paper, the output of local 2D self-attention at position $\PixelIJ$ is computed as:
\begin{equation}
y_{ij} = 
    \sum_{\mathclap{a, b \in \, \LocalWindow}}
        \texttt{softmax}_{a b}\left(  q_{i j}^\top k_{a b} + q_{i j}^\top r_{a-i,b-j} \right)  v_{a b}
\end{equation}  \label{eq:attention}
where the queries $q_{ij} = W_Q x_{ij}$, keys $k_{ab} = W_K x_{ab}$, and values $v_{ab} = W_V x_{ab}$ are linear transformations of the pixels, and $r_{a-i,b-j}$ is a learned relative position based embedding. Following the Transformer~\cite{vaswani2017attention}, we also use multihead attention, where we run multiple instances of the self-attention in parallel with different parameters. However, each head \emph{shares} the parameters for the relative embeddings $r_{a-i,b-j}$. For an attention window of size $k$ around each pixel, we factorize the relative embeddings along height and width following~\cite{ramachandran2019standalone}, and we allocate half the channels within a head to each of these. Keeping the dimension per head fixed at $16$ as mentioned in the paper, this gives a constant $2(k-1) * 16$ parameters per attention layer layer for $r_{a-i,b-j}$. In contrast, if the channels in an attention layer are $d$, then each of the three linear transformations has $d^2$ parameters. Thus the ratio of parameters in the relative embeddings as compared with the linear projections is $\frac{2(k-1) * 16}{3d^2}$, which is small for typical values of $k$ and $d$.

\begin{table}[h]
\centering
\begin{tabular}{lcc|cc}
\toprule
\multicolumn{1}{l}{\textbf{Dimension}} & \multicolumn{2}{c|}{\textbf{Values}} & \textbf{Accuracy} & \multirow{2}{*}{\textbf{\begin{tabular}[c]{@{}c@{}}Baseline \\ $\Delta$\end{tabular}}} \\& \textbf{Baseline}  & \textbf{Scaled}       &  &  \\
\midrule
Layers & 50 & 98 & 81.4 & 0.9 \\
$\VFactor$  & 1.0 & 3.0 & 81.0 & 0.5 \\
$\BaseWidthFactor$& 1.0  & 1.25 & 80.9 & 0.4 \\
$\BottleneckFactor$ & 4.0  & 6.5 & 80.6 & 0.1 \\
$\QKFactor$  & 1.0 & 6.5 & 80.3 & -0.2 \\
\bottomrule
\end{tabular}
    \caption{\textbf{Increasing the number of channels for the values and number of layers has the most impact on accuracy.}}
    \label{table:experiments:scaling}
\end{table}

\begin{table*}[h]
\centering
\begin{tabular}{c|c|c|c|c|c|c|c|c|c|c| c}
\toprule
\multirow{2}{*}{\textbf{\begin{tabular}[c]{@{}c@{}}HaloNet \\ Model \end{tabular}}} & \textbf{\BlockSize} & \textbf{\HaloSize} & \textbf{\VFactor} & \textbf{\BottleneckFactor} & \multirow{2}{*}{\textbf{\begin{tabular}[c]{@{}c@{}}Total \\ Layers \end{tabular}}} & \textbf{\ThirdGroupLayers} & \textbf{\ImageSize} & \textbf{\FinalFCWidth} &\multirow{2}{*}{\textbf{\begin{tabular}[c]{@{}c@{}}Params \\ (M)\end{tabular}}} & \multirow{2}{*}{\textbf{\begin{tabular}[c]{@{}c@{}}EfficientNet \\ Params (M)\end{tabular}}} &
\multirow{2}{*}{\textbf{\begin{tabular}[c]{@{}c@{}}EfficientNet \\ Image Size (M)\end{tabular}}}\\
 & & & & & & & & & & \\
\midrule
H0     & 8 &  3 & 1.0 & 0.5 & 50 & 7 &256 & -- & 5.5 & B0: 5.3 & 224\\
H1     & 8 & 3  & 1.0 & 1.0 & 59 & 10 & 256 & -- & 8.1 & B1: 7.8 & 240 \\
H2     & 8 & 3  & 1.0 & 1.25 & 62 & 11 & 256 & -- & 9.4 & B2: 9.2 & 260 \\
H3     & 10 & 3 & 1.0  & 1.5 & 65 & 12 & 320 & 1024 & 12.3 & B3: 12 & 300\\
H4     & 12 & 2 & 1.0  & 3 & 65  & 12 & 384 & 1280 & 19.1 & B4: 19 & 380 \\
H5     & 14 & 2 & 2.5  & 2 & 98  & 23 & 448 & 1536 & 30.7 & B5: 30 & 456\\
H6     & 8 & 4 & 3 & 2.75 & 101 & 24 & 512 & 1536 & 43.4 & B6: 43 & 528\\
H7     & 10 & 3 & 4 & 3.5 & 107 & 26 & 600 & 2048 & 67 & B7: 66 & 600\\
\bottomrule
\end{tabular}
\caption{\textbf{Configurations of HaloNet models, each of which matches a model from the EfficientNet family in terms of parameters}. The number of heads in the four stages are $(4, 8, 8, 8)$. The notations are: image size $\ImageSize$, query block size $\BlockSize$, halo size $\HaloSize$, attention output width multiplier $\VFactor$, bottleneck output width multiplier $\BottleneckFactor$, number of bottleneck blocks in the third group $\ThirdGroupLayers$, and final $1 \times 1$ conv width $\FinalFCWidth$}
    \label{tab:appendix:model-description}
\end{table*}

\section{Study of enlarging self-attention models}
\label{subsec:scaling-halonet}
In Section~\ref{sec:experiments:studies:halonet-arch}, we presented some scaling properties of our models. In Table~\ref{table:experiments:scaling}, we try to understand which other parts of our models most impact accuracy. For our study, we increase the size of HaloNet-50 by scaling different hyperparameters to reach a parameter budget of $30$ million.
We find that adding more computation in the attention by increasing $\VFactor$ and adding more layers are most fruitful scaling dimensions for increasing accuracy.

\section{Experimental details, hyperparameters}
In this section, we list the experimental details and model configurations that were omitted from the main body in interest of space

\subsection{Experimental details for model studies}
In Sections \ref{sec:experiments:studies:attention-vs-conv} and \ref{sec:experiments:studies:halonet-arch}, all the HaloNet-50 models use the same layer allocations and channels widths as the standard ResNet-50~\cite{he2016deep} model. Both ResNet-50 and HaloNet-50 models were trained for $150$ epochs on $256\times256$ size images with a learning rate of $0.1$. For the experiments with RandAugment, we used a weight decay of $0.00004$ for the settings that used RandAugment~\cite{cubuk2019randaugment}, and $0.00008$ otherwise. Using a weight decay of $0.00008$ with RandAugment seemed to have a negative impact on accuracies with ResNet-50. We used a RandAugment magnitude of $10$ in these sections. For HaloNet-50, we used a block size $\BlockSize=8$, and halo $\HaloSize=3$. We fixed the number of channels per head to be $16$. For the SASA models in section, we used a pixel \emph{centered} window of size $7\times7$ following~\cite{ramachandran2019standalone}.

\subsection{HaloNet Models}
\label{sec:detailed-models}
In Table~\ref{tab:appendix:model-description}, we describe the configurations of our HaloNet models, $H1-H7$. The hyperparameters in the HaloNet family are: image size $\ImageSize$, query block size $\BlockSize$, halo size $\HaloSize$, attention output width multiplier $\VFactor$, bottleneck output width multiplier $\BottleneckFactor$, number of bottleneck blocks in the third group $\ThirdGroupLayers$, and final $1 \times 1$ conv width $\FinalFCWidth$.
Each of our HaloNet models is trained on a comparable image size to the corresponding EfficientNet~\cite{tan2019efficientnet} model, which can be found in Table~\ref{tab:appendix:model-description}.

\subsection{Classification hyperparameters}
\label{sec:appendix:classification-hparams}
In this section we complete the details of our training and regularization setup. We used a weight decay of $2e^{-5}$ and using a cosine annealing scheme~\cite{loshchilov2016sgdr} with learning rate $0.1$. The largest models consistently overfit at the very end of training, which we attribute to the learning rate going to 0 at the end of training~\cite{yu2020bignas}. To combat this, we set the end of the cosine annealing to be $\frac{1.0}{128}$ of the original learning rate instead of $0$. For RandAugment~\cite{cubuk2019randaugment}, we grow our RangAugment magnitudes for the smallest $H0$ to the the largest $H7$ models as $6, 8, 10, 14, 17, 19, 24$ and $31$. Note that we have not extensively tuned the RandAugment magnitudes.

\subsection{Detection and instance segmentation hyperparameters}
\label{sec:detection-appendix}
We use Mask-RCNN \cite{he2017mask} for all detection and instance segmentation experiments.
We pretrain the backbone on ImageNet, mostly reusing the same hyperparameters as in Section \ref{sec:appendix:classification-hparams}.
Backbones are pretrained for $350$ epochs using an image size of $512$, which was chosen to be closer to the $1024$ image size used in detection setting.
The models were regularized with RandAug at a magnitude of $15$ and stochastic depth with probability $0.1$, and use Squeeze-Excitation with a reduction factor of $\frac{1}{8}$. 
The detection code and hyperparameters directly used the open-source TPU detection and segmentation framework
\footnote{\url{https://github.com/tensorflow/tpu/tree/master/models/official/detection}}.
During the detection / instance segmentation phase, the backbone is initialized with the pretrained weights, while the other parameters are initialized from scratch.
The model is trained for $67500$ steps with $0.1$x learning rate decays at $60000$ and $65000$ steps, uses a learning rate of $0.1$ in SGD with $0.9$ momentum, a warmup of $500$ steps with a fixed learning rate of $\frac{2}{300}$, a batch size of $64$ spread across $32$ TPUv3 cores, $1024\times 1024$ image size, an L2 weight decay of $4e^{-5}$, and multi-scale jitter with magnitudes between $[\frac{4}{5},\frac{5}{4}]$.

\section{Optimizations}
\label{sec:optimizations}
We endeavor to avoid data formatting operations whenever possible, which can slow down the model, resulting in the following two key optimizations
\begin{itemize}
    \item \textbf{Persistent blocking:} Once the image is blocked, we flatten the $(\BlockSize, \BlockSize)$ blocks to sequences of length $b^2$, and we do not reshape it back to 4D until the end of the network, implementing operations such as batch normalization~\cite{ioffe2015batch} to handle the blocked format. The image is thus processed in 5D: $(\text{Batch}, \frac{H}{\BlockSize}, \frac{W}{\BlockSize}, b^2, \ChannelSize)$ instead of $(\text{Batch}, H, W, \ChannelSize$).
    \item \textbf{Gathers with convolutions:} The haloing described in Section~\ref{sec:improved-impl} is also carried out in 5D resulting in flattened neighborhoods. For speed, we implement haloing with 3D convolutions used as gathering operations instead of slices and concatenations.
\end{itemize}

\section{ImageNet-21k Models}
\label{sec:appendix:imagenet-21k-transfer}
For our ImageNet-21k transfer experiments  Table~\ref{tab:imagenet21k-acc}), we make 4 changes to our HaloNet H4 model (See Table~\ref{tab:appendix:model-description} for specification of the H4 model). To increase the number of parameters in the model body, We increase the base width $\BaseWidthFactor$ to 2.0 (Making the base width 128, twice the normal width), and we also change $\BottleneckFactor$ from $3.0$ to the default $4.0$. We remove the final extra $1\times1$ convolution, so that the label embeddings have a large number of filters to account for the larger number of labels. Finally, we increase the number of layers in the second stage from $3$ to $4$. For the hybrid model, we use convolutions in the first two stages. 

For pretraining on $256\times256$ images, we set $\BlockSize=8$ and $\HaloSize=2$. For finetuning on $384\times384$ images, we use $\BlockSize=12$, $\HaloSize=2$, and for finetuning on $512\times512$ size images, we use $\BlockSize=16$, $\HaloSize=1$. When transferring the pretrained model, we initialize all the parameters from the pretrained checkpoint at the final step of pretraining except for the label embeddings, which are initialized to zeros, and the relative embeddings, that are initialized by linearly interpolating from the ones learned at pretraining.

\section{Impact of relative position encodings}
\cite{ramachandran2019standalone} showed that using relative position was important for achieving good accuracies. We find the same outcome with HaloNet. Using absolute factorized abosolute position encodings, which are added to the activations before local self-attention in every layer, drops accuracy from to $78.6\%$ (the first row in Table~\ref{table:experiments:component-ablation}) to $77.5\%$

\end{document}